\documentclass[letterpaper, 10 pt, conference]{ieeeconf}  

\IEEEoverridecommandlockouts                              

\usepackage{graphics} 
\usepackage{epsfig} 
\usepackage{mathptmx} 
\usepackage{times} 
\usepackage{amsmath} 
\usepackage{amssymb}  
\usepackage{booktabs}
\usepackage{multirow}
\usepackage{graphicx}
\usepackage{flushend}
\usepackage{subcaption}
\usepackage{cuted}
\usepackage{tabularx}
\usepackage{booktabs}
\usepackage{hyperref}
\usepackage{xcolor}
\usepackage{svg}

\title{\LARGE \bf
Beyond Implicit Force: Evaluating Explicit Force-Torque Proxies in Action Chunking with Transformers
}

\author{King Hang Wong, Lingqiao Liu, Feras Dayoub
\thanks{The authors are with the Australian Institute for Machine Learning (AIML), Adelaide University, Adelaide, SA 5005, Australia.
        {\tt\small \{kinghang.wong, lingqiao.liu, feras.dayoub\}@adelaide.edu.au}}%
}

\begin{document}

\maketitle

\thispagestyle{empty}
\pagestyle{empty}



\begin{abstract}

Contact-rich manipulation requires policies to infer interaction state from signals that are often weakly observable through vision and kinematics alone. Action Chunking with Transformers (ACT) has shown strong performance in fine-grained manipulation, but many deployments collect demonstrations through leader--follower teleoperation, where tracking error between commanded leader motion and executed follower motion implicitly encodes contact, resistance, and constraint violation. This paper examines whether ACT’s apparent force-awareness depends on this hidden interaction cue. We introduce an observation-centric ACT variant that predicts future follower joint states instead of leader commands, thereby removing the teleoperation-induced discrepancy signal while preserving the rest of the learning pipeline. We then evaluate whether simple joint-torque proxies, derived from onboard motor current or joint effort, can recover contact-aware behavior without external force/torque sensors. Across four real-world tasks spanning surface following, insertion, stiffness discrimination, and force-based stopping, removing the implicit cue leads to severe failures in force-critical phases. In contrast, torque-augmented policies recover robust contact behavior and improve the base ACT policy. 
These results demonstrate that, on real hardware, the implicit teleoperation cue is a recoverable source of force-awareness — where torque signals are available, a simple proxy matches, surpasses, or further enhances it.
\end{abstract}


\section{Introduction}

Robust contact-rich manipulation is challenging because crucial interaction forces, compliance, and friction are only partially observable from vision and kinematics.
Recent progress in end-to-end imitation learning has produced increasingly capable manipulation policies, ranging from large Vision--Language--Action (VLA) models to smaller Vision--Action (VA) policies. 
Action Chunking with Transformers (ACT)~\cite{alohaact} is a widely adopted VA model due to its fast single-pass inference, small weight, smooth chunked trajectories, and strong performance with as few as 20–30 demonstrations on simple tasks.

Notably, ACT’s real-world success is intertwined with how demonstrations are collected and executed. In the common leader--follower teleoperation setting used in ALOHA-style systems, the low-level controller tracks leader commands using position control; the resulting tracking discrepancies provide an \emph{implicit interaction cue} that correlates with contact onset, constraint violation, and external resistance. This cue is not an explicit force model, but acts as a latent signal that helps policies disambiguate free-space motion from high-load contact.

Crucially, this teleoperation-induced interaction cue is absent in several practical and common workflows. In kinesthetic teaching, the robot is physically guided and no separate leader arm exists; similarly, when demonstrations are collected through alternative interfaces or higher-transparency controllers that minimize tracking error, the discrepancy signal is reduced and interaction events become less visible in the recorded kinematics. 
This raises a key question: \emph{To what extent does ACT rely on this implicit teleoperation cue, and can onboard torque signals replace it on platforms lacking external force/torque sensors?}

\begin{figure}[t] 
    \centering
    \hspace*{-0.35cm}
    \includegraphics[width=1.08\columnwidth, height=7.5cm]{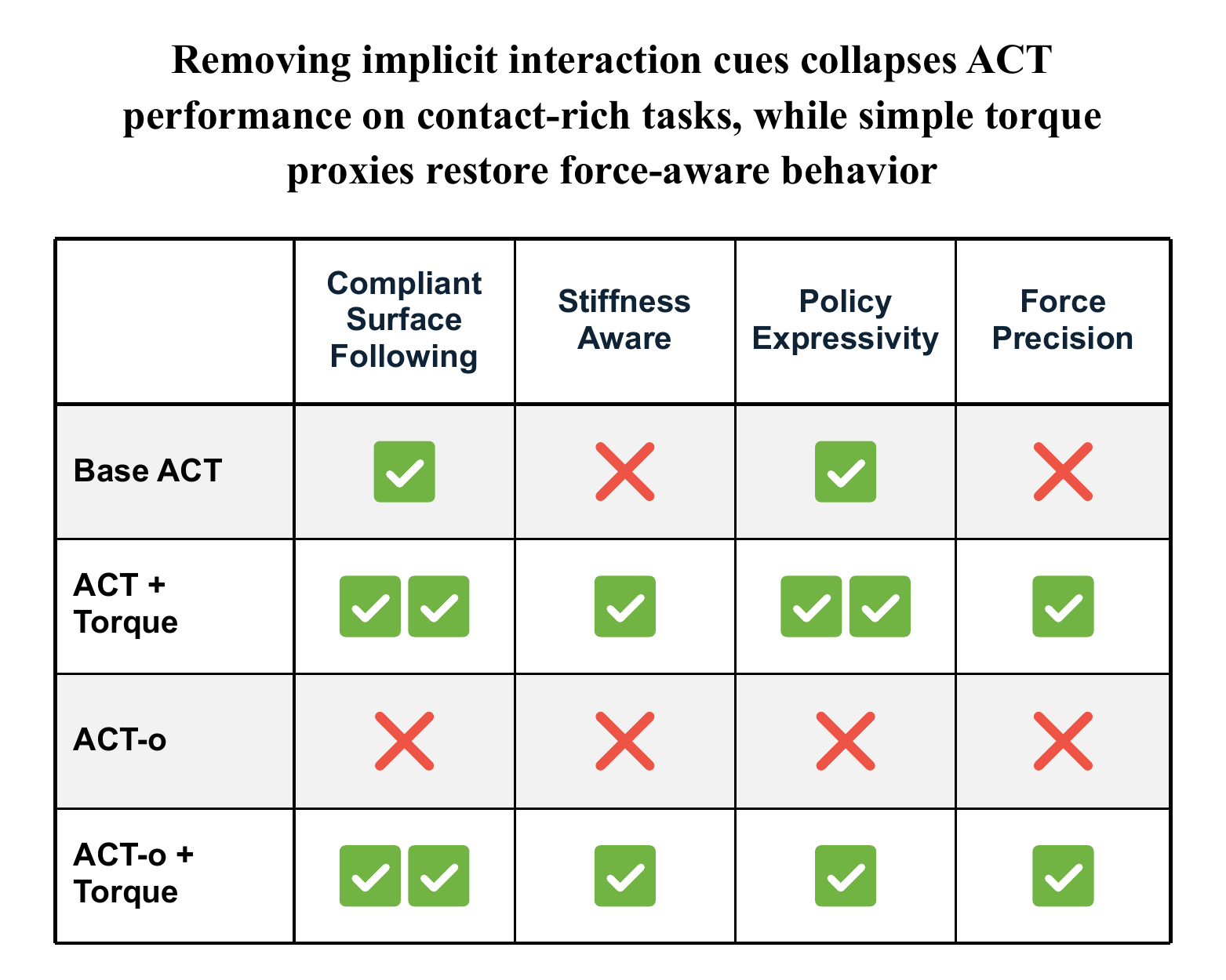}
    \caption{\textbf{Summary of model capabilities across contact-sensitive behaviors} By eliminating implicit force modeling by predicting future follower joints instead of leader joints, we collapse performance across contact-dependent behaviors \textbf{(ACT-o)}. Injecting simple force proxies (\textbf{+Torque} models) restores force-aware manipulation and enhances policy modeling.}
    \label{fig:single_column_diagram}
\end{figure}

At the same time, many research and industrial platforms lack six-axis force/torque (F/T) sensors due to cost, hardware mismatch, and integration complexity. Instead, they expose internal measurements such as motor currents or joint effort estimates, from which torque- or wrench-like proxies can be computed with varying fidelity (e.g., through filtering, compensation, or Jacobian-based mappings). Despite growing interest in multimodal sensing, there remains limited systematic evidence on how such \emph{explicit torque proxies} compare to teleoperation-induced implicit cues within ACT-style policies, and which proxy representations are most effective under realistic data scales and contact conditions.

In this paper, we present a controlled real-world study of implicit versus explicit interaction signals in ACT for contact-rich manipulation. We introduce an \emph{observation-centric} ACT formulation that predicts future follower robot states rather than leader commands, thereby removing the teleoperation-induced discrepancy cue.
We then inject explicit torque-based proxies derived from onboard measurements and evaluate whether they can restore contact-aware behavior. Across multiple real-robot contact-sensitive tasks spanning impact-like contacts, sustained frictional interaction, and state-dependent force transitions, we show that removing the implicit cue can cause catastrophic failure in force-critical phases, and that appropriately processed torque proxies can recover and enhance performance.

Our contributions are:
\begin{itemize}
    \item \textbf{Isolation of implicit interaction cues in ACT:} We formalize and ablate the teleoperation-induced discrepancy signal by introducing an observation-centric ACT target, and show that removing this cue can collapse contact-rich task performance.
    \item \textbf{Explicit torque proxy evaluation:} Along with Base ACT, we evaluate three proposed model variants and demonstrate that explicit proxies can recover contact-aware behavior in the absence of external F/T sensors.
    \item \textbf{Task-level analysis under contact:} We provide quantitative results and diagnostic analyses that characterize reactivity and stability under contact across diverse contact regimes, and we report practical guidelines for integrating torque proxies into ACT-style policies.
\end{itemize}



\section{Related Work}

Recent end-to-end imitation learning for manipulation spans large Vision--Language--Action (VLA) models and high-frequency Vision--Action (VA) policies. VLAs have demonstrated impressive generality through large-scale pretraining and broad task coverage \cite{kim2024openvla,black2024pi0,gr00tn1_2025}, but their deployment can involve higher inference overhead and lower-rate action abstractions that may be less suitable for contact-critical phases requiring rapid feedback and precise regulation. In contrast, VA policies trained on task-specific demonstrations remain a standard choice for real-robot precision control, including ACT \cite{alohaact} and Diffusion Policy \cite{chi2023diffusion}. These approaches leverage temporal structure (e.g., action chunking \cite{alohaact} or diffusion over action sequences \cite{chi2023diffusion}) to improve stability and mitigate compounding errors, and they provide strong baselines for controlled studies in contact-rich manipulation.

A long-standing line of work has shown that force/torque (F/T) feedback improves robustness in contact-rich tasks by providing direct evidence of interaction state when vision and kinematics are ambiguous. Recent few-shot imitation-learning methods combine force feedback with learned representations to enable adaptive contact-rich behaviors under limited data \cite{adaptivewiping}. In parallel, emerging work on torque-aware and force-aware large models studies how to represent and fuse interaction signals within VLA-style policies \cite{zhang2025tavla,yuforcevla}. While these results highlight the value of explicit interaction sensing, they typically assume either dedicated sensing hardware or large-scale model/data regimes that differ substantially from low-data ACT-style deployments on widely used research platforms.

Bilateral control and haptic teleoperation provide a practical pathway for collecting demonstrations and regulating contact by coupling human intent with robot interaction dynamics. Recent extensions incorporate bilateral control into ACT-style learning for contact-rich manipulation \cite{biact,kobayashi2025bilat}, and related work studies variable compliance control using haptic feedback teleoperation \cite{compact}. These systems underscore that contact performance can be influenced not only by the policy architecture but also by the data-collection and control interface, motivating further studies for careful isolation of interaction-related signals when comparing learning-based methods. Unlike these works, our work studies contact cue implicitly created by leader–follower tracking error and tests whether onboard torque proxies can replace or enhance it in non-bilateral demonstration settings.

Despite growing interest in multimodal sensing, many platforms lack dedicated six-axis F/T sensors due to cost and integration complexity. New sensor designs aim to reduce this barrier (e.g., compact 6-axis F/T sensors) \cite{coinft,umift}, but a substantial fraction of platforms still rely on onboard measurements such as motor currents or joint effort estimates. This motivates using torque- and wrench-like proxies derived from internal dynamics as a lightweight alternative for contact awareness. Our work targets this practical setting and focuses on ACT-style policies trained from limited demonstrations.

The prior work establishes strong VA baselines for precise manipulation, demonstrates the utility of explicit interaction sensing for contact-rich behaviors, and explores torque, force, and tactile representations in larger model families. However, there remains limited systematic evidence in ACT-style pipelines on (i) how performance depends on interaction cues tied to teleoperation and control interfaces, and (ii) whether explicit torque proxies derived from onboard measurements can reliably substitute for such cues when external F/T sensing is unavailable. This paper addresses these questions through controlled ablations and a task-level empirical study on contact-sensitive manipulation.


\section{Preliminaries}
\label{sec:Preliminaries}

\subsection{Teleoperation Data and the Interaction Signal Gap}
Imitation learning performance is bounded by demonstration fidelity, where control discrepancies heavily impact how models like ACT learn contact dynamics.

\begin{figure}[t] 
    \centering
    \includegraphics[width=1.0\columnwidth]{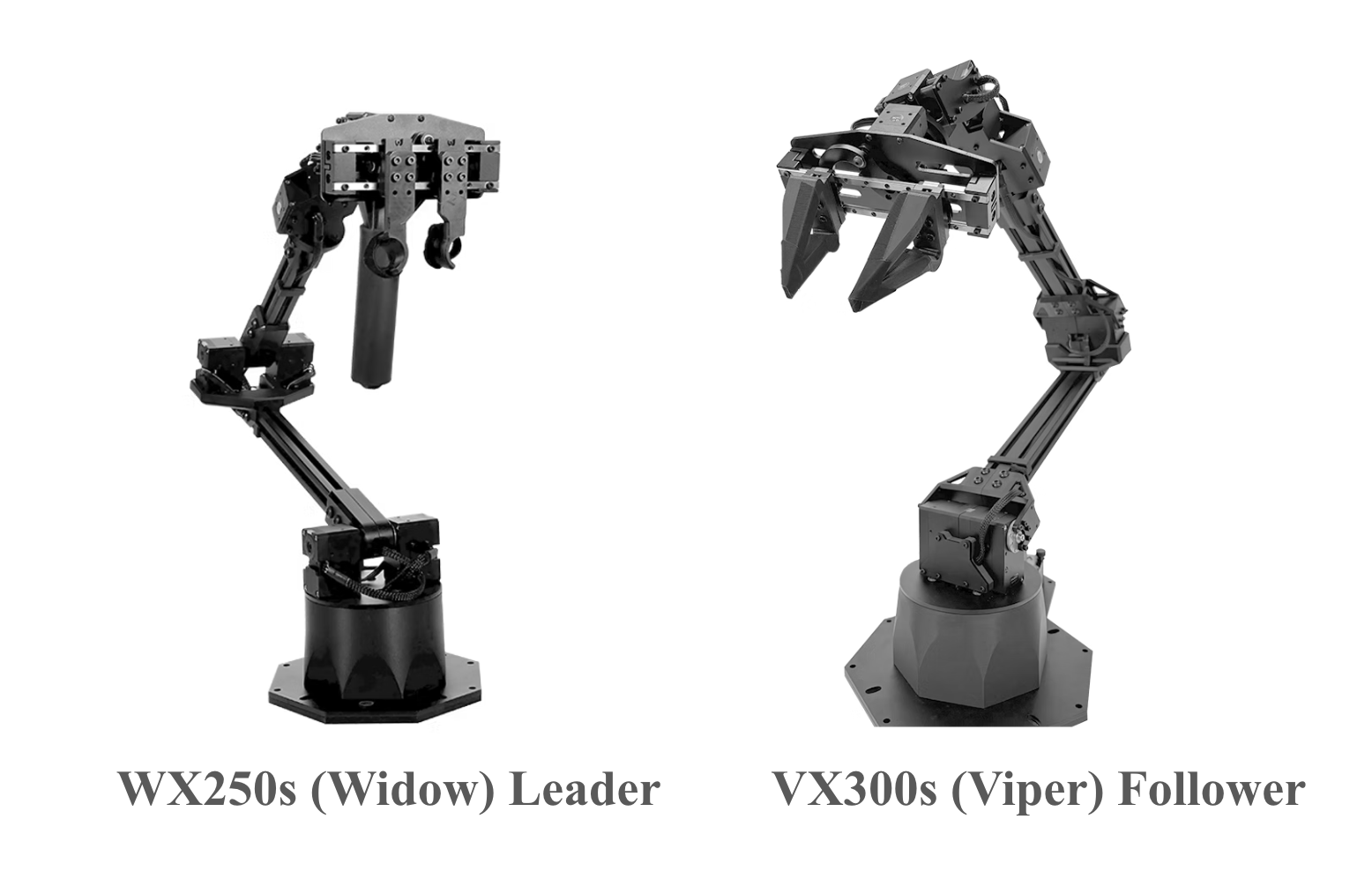}
    \caption{The ALOHA Solo teleoperating system \cite{alohaact}\cite{trossen_viperx_2024} uses a kinematically identical Leader-Follower arm pair, enabling one-to-one direct joint state mapping.}
    \label{fig:aloha}
\end{figure}

\textbf{Leader--Follower Teleoperation:} Popularized by platforms like ALOHA \cite{alohaact}, \cite{fu2024alohamobile}, \cite{zhao2025aloha}, this involves using kinematically identical "leader" arms to drive follower arms (Fig. \ref{fig:aloha}). While this joint-space mapping provides high-bandwidth coordination, it importantly introduces a latent "interaction cue": the positional discrepancy between the leader and the stalled follower during contact. Our study investigates an observation-centric ACT variant that predicts future follower states directly, thereby removing this implicit signal and exposing the model's reliance on these artifacts.

\textbf{Kinesthetic Teaching and High-Transparency Interfaces:} These respectively involve physically guiding the robot and using high accuracy controllers that either directly execute action commands or minimize tracking error. Consequently, the teleoperation-induced discrepancy signal is reduced or eliminated, making interaction events less visible in recorded kinematics.

\textbf{Impact on Policy Learning:} This hardware dependent signal gap raises a fundamental question: \emph{to what extent does ACT’s contact performance rely on the implicit interaction cue present in leader--follower teleoperation, and can it be replaced by signals available on platforms without external force/torque sensors?} Our results demonstrate that by removing implicit force modeling, the explicit re-integration of F/T data or proxies not only restores but enhances the ability to complete contact-rich tasks. They suggest explicit force modeling is not just beneficial for general policy modeling, but important for robust force-aware tasks.

\subsection{Action Chunking with Transformers (ACT)}
The ACT framework addresses the challenges of high-precision imitation learning by formulating robot control as a generative modeling task. Unlike traditional step-by-step behavior cloning, which is prone to compounding errors and drifts, ACT predicts a sequence (or "chunk") of $k$ future actions simultaneously given the current observations. The model effectively handles multi-modal action distributions while ensuring smooth, coherent trajectories even in complex tasks.

The architecture is built upon a Conditional Variational Autoencoder (CVAE). During training, an encoder $E_\phi$ maps the current state and the target action sequence to a latent variable $z$, which represents the "style" or "intent" of the motion. A Transformer-based decoder $D_\theta$ then reconstructs the action chunk conditioned on this latent $z$ and the current proprioceptive state. The training objective is to minimize a loss function consisting of a reconstruction term and a KL-divergence regularization term:

\begin{equation}
\begin{split}
\mathcal{L}_{ACT} = \mathbb{E}_{z \sim Q} [ \| a_{t:t+n} - \hat{a}_{t:t+n} \|_2^2 ] \\
+ \beta D_{KL}(Q(z|s_t, a_{t:t+n}) \| P(z|s_t))
\end{split}
\end{equation}

where $a_{t:t+n}$ is the target action sequence from the sampled training data, $\hat{a}_{t:t+}$ is the predicted chunk, and $\beta$ is a hyperparameter balancing the two terms.

\textbf{Implicit Force via Joint Discrepancy} In standard ACT implementations, the policy $\pi$ typically maps observations $O$ to a sequence of leader joint positions $A = \{q_{\text{leader}, t}, \dots, q_{\text{leader}, t+n}\}$. This conventional setup relies on the inherent discrepancy $\delta = q_{\text{leader}} - q_{\text{follower}}$, which captures both the intended motion and a latent, implicit force-awareness generated by the operator's physical response to resistance.

\subsection{The ALOHA Platform and ViperX Dynamics}
Our experimental setup utilizes the ALOHA Solo platform, specifically configured with a Trossen Robotics WidowX-250s and ViperX-300s Leader-Follower arm pair \cite{trossen_viperx_2024}. Each arm features 6 Degrees of Freedom (DoF), and the paired-arms are kinematically identical. The ViperX Follower actuators provide real-time feedback for joint position, velocity, and effort (current), which we leverage as torque proxies.

In contact-rich manipulation, the relationship between the joint torques $\tau$ and the external wrench $F_{ext}$ applied at the end-effector is governed by the manipulator Jacobian $J(q)$:
\begin{equation}
\label{eq:jacobian}
\tau = J(q)^T F_{ext} + \tau_{dyn}(q, \dot{q}, \dots)
\end{equation}
where $\tau_{dyn}$ accounts for the internal dynamics of the arm, including gravity, Coriolis forces, and friction. 
Classical contact-rich control relies on the manipulator Jacobian to map joint torques to task-space wrenches, requiring an explicit analytical dynamics model ($\tau_{\text{dyn}}$) to isolate external forces from gravity and inertia. Our empirical results demonstrate that raw, joint-level effort signals provide a highly robust alternative foundation for ACT. Furthermore, our results demonstrate that calculating the full residual dynamics term $\tau_{dyn}$ is unnecessary for an effective torque-augmented policy. 



\section{Methodology}

\subsection{Observation-Centric Prediction and Force Decoupling}

By substituting observation joint states for leader-arm states as the prediction target, we decouple the ACT architecture from implicit teleoperation dynamics. This shift more strictly constrains actions to valid state spaces, and better supports wider teleoperation setups discussed in Section \ref{sec:Preliminaries}.

\textbf{Joint Target Reformulation} 
We redefine the decoder prediction target as the follower's future proprioceptive states \(Q \triangleq Q_{t:t+n} = \{q_{\mathrm{follower},\,t+1}, \ldots, q_{\mathrm{follower},\,t+n}\}\). This reformulation allows us to isolate the policy from the leader arm's mechanical artifacts and digitally configured compliance, and experimentally evaluate the performance impact when the model is deprived of these latent force cues. The reconstruction loss $\mathcal{L}_{\text{rec}}$ is thus reformulated to minimize the error between the predicted and actually achieved robot configurations:

\begin{equation}
\mathcal{L}_{\text{rec}} = \mathbb{E}_{z, O, Q} \left[ \| Q - \hat{Q}(z, O) \|_2^2 \right]
\end{equation}

where $z$ is the latent variable sampled from the CVAE encoder.

\subsection{Explicit Force Proxies via Servo Current}
The removal of leader actions eliminates the implicit force modeling naturally embedded in the leader-follower lag. To recover this information, we propose ablative variants to introduce explicit force proxies using screw-axis torque signals $\tau$. The integration of $\tau$ allows the transformer to learn a mapping between torque transients and physical resistance. This enables task completion without the need for an analytical gravity or friction compensation model $G(q)$. The dynamics are governed by:

\begin{equation}
\label{eq:dynamics}
\tau_{\text{total}} = M(q)\ddot{q} + C(q, \dot{q})\dot{q} + G(q) + \tau_{\text{ext}}
\end{equation}

where in quasi-static states, as $|\ddot{q}|, |\dot{q}| \rightarrow 0$ (i.e., pushing against a fixed object), only $G(q), \tau_{ext}$ remain.

By providing $\tau_{\text{total}}$ as a direct input, a torque-aware policy learns to compensate for $\tau_{\text{ext}}$ (external contact forces) and $G(q)$ implicitly via its learned modes. Experimental results in Section \ref{sec:Results} demonstrate that this approach is robust in various contact-rich tasks, where  effort change patterns serve as an informative proxy.

\textbf{Injection Method} To integrate this additional modality into the ACT architecture, we treat torque proxies as "first-class" inputs alongside proprioceptive joint states. We found that directly concatenating raw effort values with their corresponding joint states—prior to hidden-dimension projection—allows the model to utilize a \textit{shared joint-state embedding} that projects the 14 (joints + effort)-dimensional input into the hidden-layer dimension: $\mathbb{R}^{14} \to \mathbb{R}^{d_{hidden}}$. Empirical results indicate the embedding's strong efficacy in both the CVAE encoder and the transformer encoder.

\begin{figure}[t] 
    \centering
    \includegraphics[width=1.0\columnwidth]{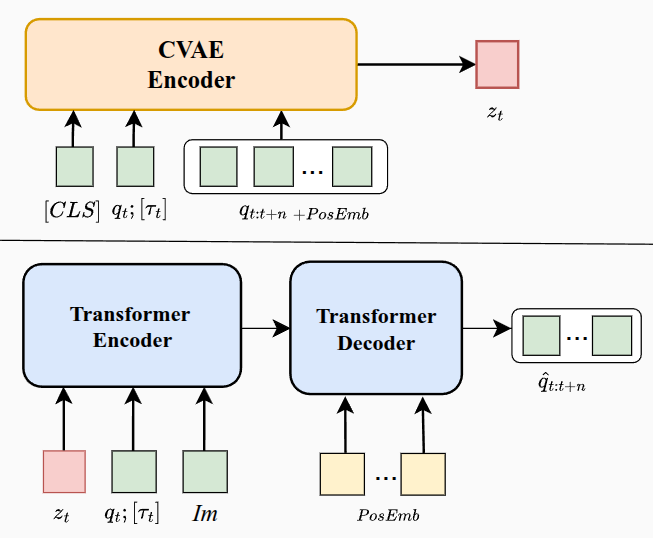}
    \caption{\textbf{Overview of the torque-augmented ACT architecture and task performance.} Standard leader-follower teleoperation embeds implicit interaction cues that Base ACT relies upon for contact tasks. Isolating the policy from these latent signals (\textit{ACT-o}) results in a systemic performance collapse. Then, treating screw-axis torque $\tau$ as a first-class input alongside joint states $q_t$, our modified CVAE (top) and Transformer architecture (bottom) restores force-aware behavior and improves stability across contact-rich manipulation tasks.}
    \label{fig:arch}
\end{figure}

\begin{table}[h]
\caption{Comparison of Prediction Targets, Input Modalities, and Dynamics Modeling}
\label{table:comparison}
\centering
\begin{tabular}{l|cc}
\hline
\textbf{Feature} & \textbf{Standard ACT} & \textbf{Proposed (Ours)} \\ \hline
Prediction Target & Action Joints ($a_t$) & Observation Joints ($q_t$) \\
Force Modeling & Implicit (lag-based) & Explicit (screw-axis $\tau$) \\ \hline
\end{tabular}
\end{table}

\subsection{CVAE Role and Input Reformulation}

To accommodate this shift, the CVAE encoder is repurposed to capture physically aware robot trajectory modes rather than human control artifacts. While the original ACT encoder ingests positionally-encoded leader-arm actions $A_{t:t+n}$, our modified encoder processes the future follower states $Q_{t:t+n}$. To evaluate the impact of explicit haptic feedback, we introduce an ablative variant where the encoder also ingests screw-axis torque proxies $\tau$. 

In the augmented variant, the encoder produces the latent $z = E(q_t, Q_{t:t+n}, \tau_{t})$ (Fig. \ref{fig:arch}, top). This allows us to compare a model that purely reconstructs state-space trajectories against one where $z$ explicitly characterizes environmental interactions and loading conditions. The latent distribution is optimized via the variational objective:

\begin{equation}
\mathcal{L}_{\text{CVAE}} = \mathcal{L}_{\text{rec}} + \beta D_{KL}[q(z|Q, [\tau]) \| p(z)]
\end{equation}

where $[\tau]$ denotes the optional inclusion of torque data depending on the model variant being evaluated.

\section{Experiments}

\subsection{Hardware Setup}

We conduct all experiments on a physical ALOHA Solo teleoperation system to ensure the fidelity of contact dynamics, which are notoriously difficult to bridge from simulation to reality. The hardware stack consists of an Interbotix ViperX-300s Follower arm and a WidowX-250s Leader arm, where the Follower is solely responsible for any object interaction. For visual input, the system utilizes one high-angle BEV camera, one low-profile side camera, and one wrist-mounted camera on the Follower end-effector.

Data collection and policy inference are synchronized at 50~Hz across all modalities to capture varied torque behaviors without the overhead of high-precision servoing. Demonstrations and Inference control is performed entirely in joint space. We estimate joint torque $\tau$ via the linear relationship between motor current $I$ and the manufacturer-defined torque constant $k_t$, such that $\tau = k_t I$.

Model training is distributed across NVIDIA RTX 4090, A6000, and A5000 GPUs. Real-time inference is standardized on a local RTX 4090 to maintain a consistent control loop.

\subsection{Model Architecture}
To evaluate the impact of different sensory and control modalities on contact-rich manipulation, we compare four variations of the Action Chunking Transformer (ACT) architecture as summarized in Table~\ref{tab:model_variants} below.

\begin{table}[ht]
\centering
\caption{ACT Model Variants and Force-Awareness}
\label{tab:model_variants}
\footnotesize
\begin{tabularx}{\columnwidth}{l l X}
\toprule
\textbf{Variant} & \textbf{Target} & \textbf{Force Mechanism} \\ \midrule
Base ACT & Leader ($A$) & Implicit (via discrepancy $\delta$) \\ \addlinespace
ACT-o & Follower ($Q$) & None (Implicit $\delta$ removed) \\ \addlinespace
ACT + $\tau$ & Leader ($A$) & Hybrid (Implicit $\delta$ + Explicit $\tau$) \\ \addlinespace
ACT-o + $\tau$ & Follower ($Q$) & Explicit (Effort $\tau$ only) \\ \bottomrule
\end{tabularx}
\end{table}

\subsection{Contact-Dynamic Task Design}
\label{sec:contact-task-design}
The following \textbf{contact dynamics}-focused task design principles categorize our experimental evaluation:

\textbf{Surface Following and Normal Force}
We examine the model's ability to maintain a consistent normal force during dynamic lateral movement. We test whether the torque proxy can stabilize the (joint-space) policy against varying resistance levels encountered when moving across a surface, preventing the "hovering" failure mode common in purely kinematic policies.

\textbf{Multi-mode, Precision, and Long Horizon}
This testing regime focuses on whether a model can model multi-mode, distinct task phases while injecting torque as a primary driving signal. We investigate if increased expressiveness provided by the torque signal enables robust policy learning without introducing drift or instability.

\textbf{Stiffness \& Reactivity}
We test each model’s response to varying material stiffness. This tests whether the policy modulates its behavior based on instantaneous mechanical feedback rather than relying on visual deformation. Given visually identical objects, we evaluate if the model can recognize a rapid torque spike patterns to create distinct observation state mappings.

\subsection{Task Metrics}

\textbf{Task Success:} 
Following prior work \cite{compact, biact, kobayashi2025bilat} we use task success as the primary evaluator. This is defined as the binary completion of a task within a predefined time horizon, verified through visual alignment with a goal state. 

\textbf{Latent Observations:} 
For assessing visually indistinguishable states (Fig. \ref{fig:foam_press_main}), we leverage an end-effector (EEF) wrench $\mathcal{F}_{ext}$ proxy via torque values. This is derived using the analytic Jacobian $J(\theta)$ and measured joint torques $\tau$ through Eq.~\ref{eq:jacobian},~\ref{eq:dynamics} under the quasi-static condition assumption.

\subsection{Task Setup}

We design a suite of common manipulation tasks to evaluate force-torque signal behaviors. To enhance data efficiency and policy robustness, we implement $SE(2)$ object start pose randomization within a constrained interaction region. These are tracked via overhead overlay to ensure comprehensive sampling and in-distribution evaluation. We validated our hyperparameter configurations (Table~\ref{table:hyperparameters}) by confirming that at least one model variant consistently generates visually successful rollouts for each task. 

\begin{table}[t]
\centering
\caption{Hyperparameter Configuration for Model Training}
\label{table:hyperparameters}
\begin{tabular}{lr}
\hline
\textbf{Hyperparameter} & \textbf{Value} \\ \hline
Training Epochs         & 8,000          \\
Chunk Size              & 100            \\
Batch Size              & 16             \\
Learning Rate           & $2 \times 10^{-5}$ \\
Feedforward Dimension ($d_{ff}$) & 3,200 \\
Hidden Dimension ($d_{hidden}$) & 512 \\
KL Weight ($\beta$)     & 10             \\ \hline
\end{tabular}
\end{table}

\subsubsection{\textbf{Board Wipe- Surface Following \& Normal Force}}

Board Wipe (Fig. \ref{fig:board_wipe_main}) shifts the focus to surface-following behavior, where the agent must generate a downward force $F_x$ while executing lateral wiping. A successful trial is defined by the complete removal of the target marking or the presence of visible streaking along the tool path, indicating consistent surface contact.

\subsubsection{\textbf{Plug Insert- Multi-Mode, Precision}}

Plug Insert (Fig. \ref{fig:plug_insert_main}) is a high-precision, multi-mode manipulation challenge requiring both alignment and force control. The robot first corrects a tilted plug resting on a powerboard before executing a vertical press to complete the insertion. We test whether torque proxies enable successful insertion and whether they introduce adverse policy drift.

\subsubsection{\textbf{Soft Bottle Press- Stiffness Sensitivity}}

In Soft Bottle Press (Fig. \ref{fig:bottle_main}) we vertically squash a collapsible bottle with two mechanical states: High-Stiffness (Locked) and Low-Stiffness (Vented). The high-stiffness state produces a resistive torque spike upon compression. A force-aware model must recognize these as a visually similar but mechanically distinct state, triggering the exit action if High-Stiffness.

\subsubsection{\textbf{Foam Stop- Force-Equilibrium Behavior}}
In Foam Stop (Fig. \ref{fig:foam_press_main}) the end-effector is sent towards a textureless block of compliant foam with an unclear geometric outline, and is demonstrated to stop after we observe light resistance while maintaining EEF verticality. We vary the height of the textureless foam to test the model's ability to learn stopping modes based on force feedback. Although an uncalibrated side-view camera provides some visual context, the lack of distinct surface features and varying heights limits its informativeness.
%


\begin{table*}[t]
\centering
\begin{tabular}{@{}l ccc cc c@{}}
\toprule
\multirow{2}{*}{\textbf{Configuration}} & \multicolumn{3}{c}{\textbf{Soft Bottle Press ($n_{stiffmode}=10$)}} & \multicolumn{2}{c}{\textbf{Plug Insertion ($n=10$)}} & \textbf{Board Wipe ($n=20$)} \\
\cmidrule(lr){2-4} \cmidrule(lr){5-6} \cmidrule(lr){7-7}
 & Overall & Low-Stiffness & High-Stiffness & 1-Step & 1 OR 2-Steps & Overall \\
\midrule
Base ACT        & 50\%           & \textbf{100\%} & 0\%            & 20\%           & 30\%           & 60\% \\
ACT + Torque    & \underline{70\%} & 60\%           & \textbf{80\%}  & \textbf{40\%}  & \textbf{70\%}  & \underline{90\%} \\
ACT-o           & 25\%           & 50\%           & 0\%            & 0\%            & 0\%            & 15\% \\
ACT-o + Torque  & \textbf{75\%}  & \underline{80\%} & \underline{70\%} & \underline{10\%} & \underline{50\%} & \textbf{95\%} \\
\bottomrule
\end{tabular}
\caption{\textbf{Performance success rates across robotic tasks and model configurations.} Results show success percentages for \textbf{Soft Bottle Press} ($n_{total}=80$), \textbf{Plug Insertion} ($n_{total}=40$), and \textbf{Board Wiping }($n_{total}=80$). For \textbf{Plug Insertion}, a "rollout" is defined as either a single successful insertion or a pair of consecutive attempts; if the first attempt contacts the plug but fails to insert, a second attempt is permitted and the two are counted as a single rollout unit.}
\label{tab:performance_table}
\end{table*}

\begin{figure*}[t]
    \centering
    \includegraphics[width=1.0\textwidth]{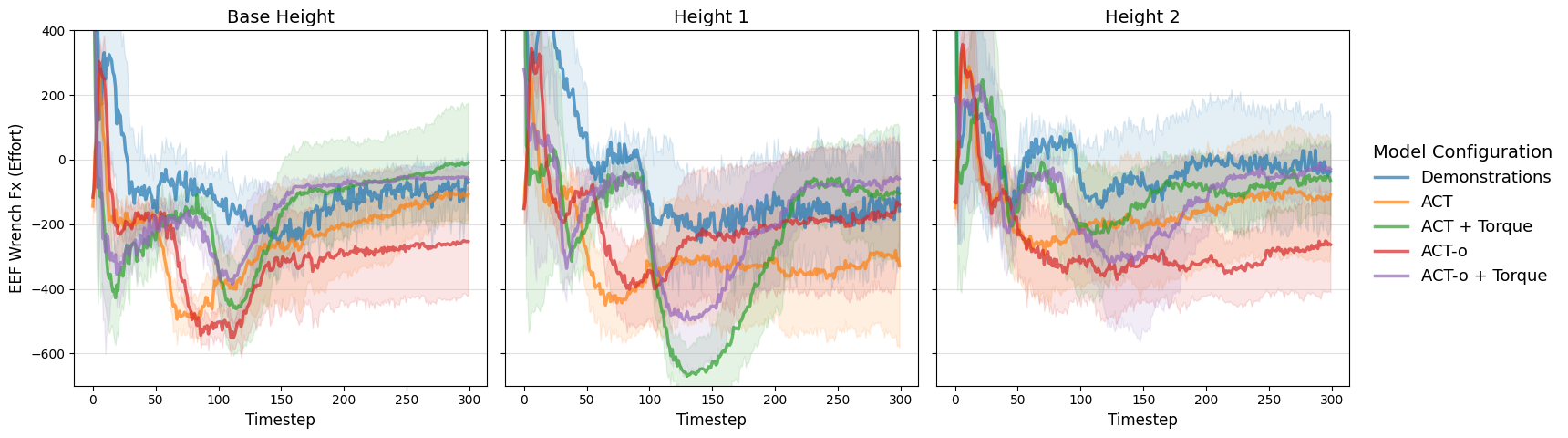}
    \caption{\textbf{Foam Stop Equilibrium Convergence} End-effector $F_x$ (in units of effort) values averaged over all \textbf{Foam Stop} (Fig. \ref{fig:foam_press_main}) episodes, grouped by the model variant at $n_{(config, mode)} = 10$ rollouts, with $n_{total}=120$ rollouts. This figure illustrates how closely different ACT-variants mimic the demonstration as the end-effector approaches an equilibrium towards the end of an episode. The demonstration, generated from averaged human demonstrations, settles into a steady, non-zero value.}
    \label{fig:torque_plot}
\end{figure*}

\section{Results}
\label{sec:Results}
\subsection{Results Overview}
We summarize critical findings based on design principles in \ref{sec:contact-task-design}.
While our methodology encompasses four distinct tasks, only three are represented in Table \ref{tab:performance_table}; the Foam Stop task is evaluated separately via a force-profile plot in Fig. \ref{fig:torque_plot} to capture its latent dynamic response.

\begin{itemize}
    \item \textbf{Impact of Removing Leader Joint Targets:} Removing leader-joint supervision causes substantial degradation in contact-dependent behavior. In the absence of explicit torque proxies, \textbf{ACT-o} often produces hesitant, stalled, or incomplete trajectories and fails entirely in several task conditions.
    
    \item \textbf{Force as a Geometric Anchor:} In \textit{Board Wipe} and \textit{Foam Stop}, torque proxies mitigated ``height-stopping'' issues, enhancing both \textbf{Base ACT} and \textbf{ACT-o}.
    
    \item \textbf{Disambiguating Latent States:} Torque feedback allows the \textit{Soft Bottle Press} policy to distinguish between visually identical but mechanically distinct states (e.g., Low vs. High Stiffness).
    
    \item \textbf{Precision and General Behavior Learning:} In \textit{Plug Insert}, torque-aware configurations exhibit more frequent, gentler ``tap-like'' adjustments, resulting in superior positional accuracy.
\end{itemize}

\begin{figure*}[p] 
    \centering

    \begin{subfigure}[b]{0.24\textwidth}
        \centering
        \includegraphics[width=\textwidth]{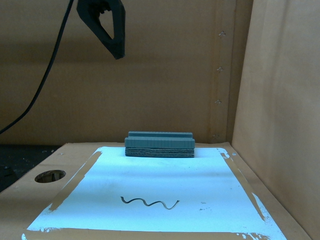}
        \caption{Approach Eraser}
        \label{fig:wipe_a}
    \end{subfigure}
    \hfill
    \begin{subfigure}[b]{0.24\textwidth}
        \centering
        \includegraphics[width=\textwidth]{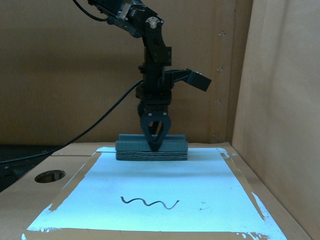}
        \caption{Grasp Eraser and Lift}
        \label{fig:wipe_b}
    \end{subfigure}
    \hfill
    \begin{subfigure}[b]{0.24\textwidth}
        \centering
        \includegraphics[width=\textwidth]{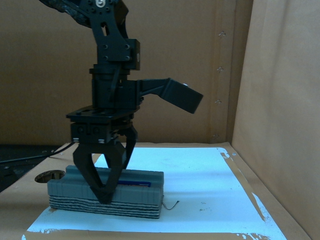}
        \caption{Apply Downward Force}
        \label{fig:wipe_c}
    \end{subfigure}
    \hfill
    \begin{subfigure}[b]{0.24\textwidth}
        \centering
        \includegraphics[width=\textwidth]{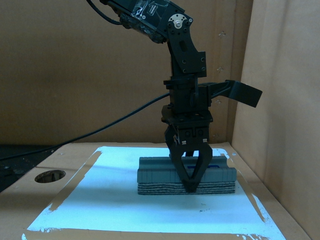}
        \caption{Perpendicular Motion}
        \label{fig:wipe_d}
    \end{subfigure}
    \caption{\textbf{Board Wipe} This task involves (a) initial approach, (b) establishing a stable grasp, then lifting and repositioning the eraser in free space, (c) applying downward force, and (d) performing a clear, strong wiping motion with appropriate vertical force while moving orthogonally across the board.}
    \label{fig:board_wipe_main}

    \vspace{1.0em} 

    \begin{subfigure}[b]{0.24\textwidth}
        \centering
        \includegraphics[width=\textwidth]{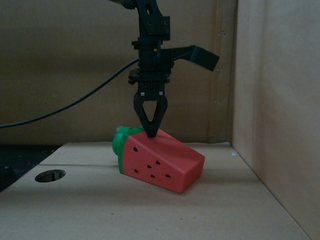}
        \caption{Approach Bottle}
        \label{fig:bottle_a}
    \end{subfigure}
    \hfill
    \begin{subfigure}[b]{0.24\textwidth}
        \centering
        \includegraphics[width=\textwidth]{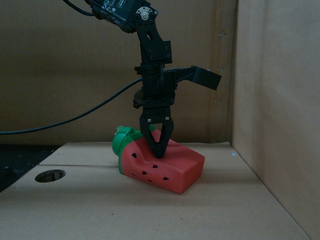}
        \caption{Initial Contact}
        \label{fig:bottle_b}
    \end{subfigure}
    \hfill
    \begin{subfigure}[b]{0.24\textwidth}
        \centering
        \includegraphics[width=\textwidth]{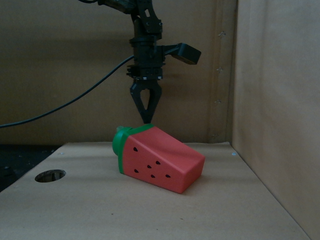}
        \caption{High Stiffness}
        \label{fig:bottle_c}
    \end{subfigure}
    \hfill
    \begin{subfigure}[b]{0.24\textwidth}
        \centering
        \includegraphics[width=\textwidth]{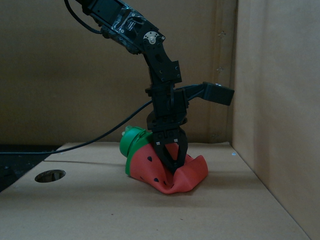}
        \caption{Low Stiffness}
        \label{fig:bottle_d}
    \end{subfigure}
    \caption{\textbf{Soft Bottle Press} The robot performs a controlled indentation or recovery task on a deformable, watermelon-shaped soft collapsible water bottle. The gripper first approaches the bottle then presses against the bottle, allowing it to process the resistance via torque signals. Depending on the inferred stiffness mode, the robot either retreats as in Fig. 6(c), or it continues pushing to collapse the bottle as in Fig. 6(d).}
    \label{fig:bottle_main}

    \vspace{1.0em}

    \begin{subfigure}[b]{0.24\textwidth}
        \centering
        \includegraphics[width=\textwidth]{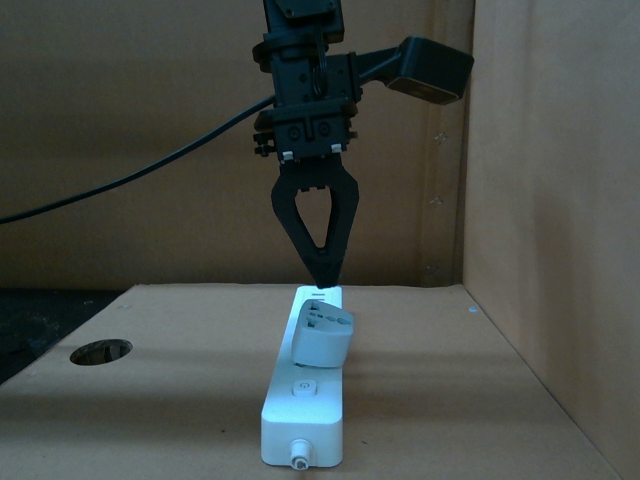}
        \caption{Identify Tilt}
        \label{fig:plug_a}
    \end{subfigure}
    \hfill
    \begin{subfigure}[b]{0.24\textwidth}
        \centering
        \includegraphics[width=\textwidth]{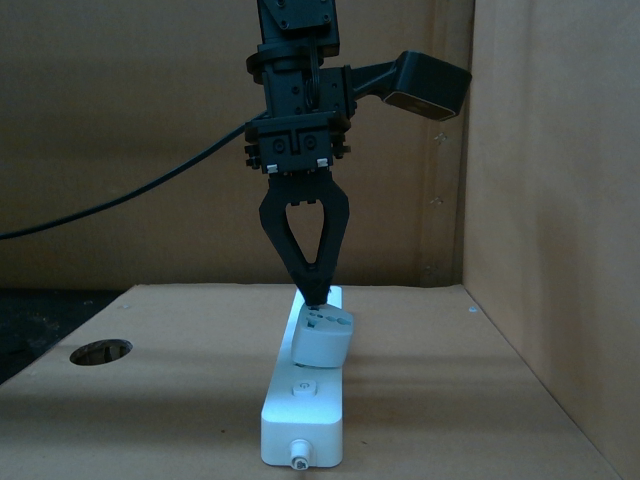}
        \caption{Adjustment Motion}
        \label{fig:plug_b}
    \end{subfigure}
    \hfill
    \begin{subfigure}[b]{0.24\textwidth}
        \centering
        \includegraphics[width=\textwidth]{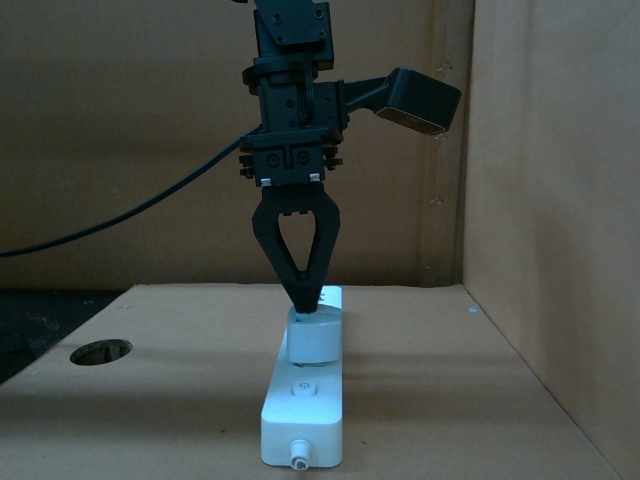}
        \caption{Align with Upward Plug}
        \label{fig:plug_c}
    \end{subfigure}
    \hfill
    \begin{subfigure}[b]{0.24\textwidth}
        \centering
        \includegraphics[width=\textwidth]{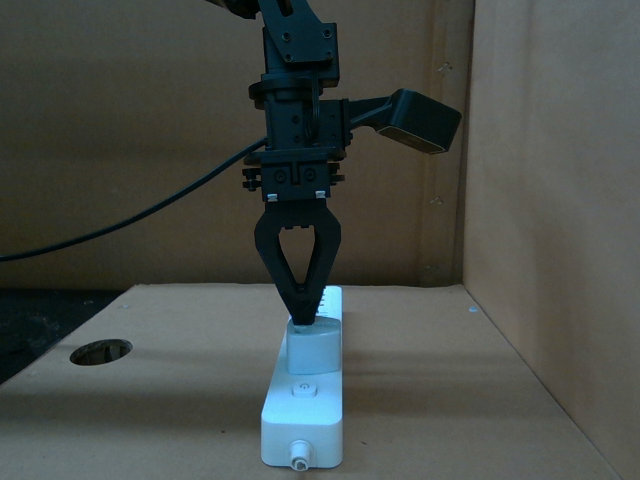}
        \caption{Apply Downward Force}
        \label{fig:plug_d}
    \end{subfigure}
    \caption{\textbf{Plug Insertion} The robot needs to perform a precise tilt-adjustment and plugging operation into a power strip. The process begins with the initial approach and alignment over the target socket. The end-effector is demonstrated to lightly push the plug to adjust the tilt until it correctly faces vertically. Finally, the system executes the insertion by applying controlled downward force to overcome friction, ensuring a secure mechanical connection.}
    \label{fig:plug_insert_main}

    \vspace{1.0em}

    \begin{subfigure}[b]{0.24\textwidth}
        \centering
        \includegraphics[width=\textwidth]{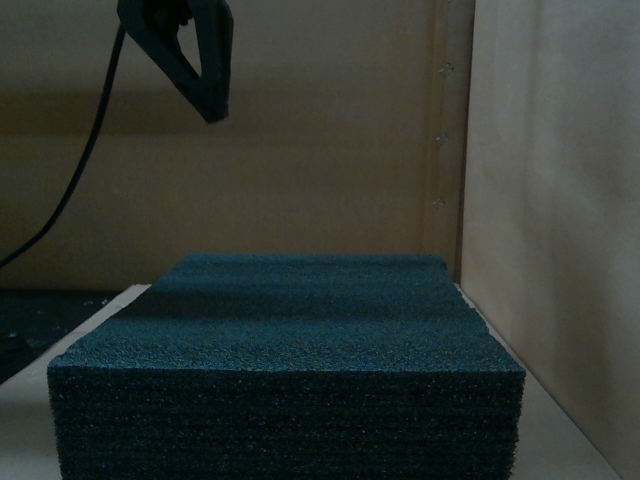}
        \caption{Approach Foam}
        \label{fig:foam_a}
    \end{subfigure}
    \hfill
    \begin{subfigure}[b]{0.24\textwidth}
        \centering
        \includegraphics[width=\textwidth]{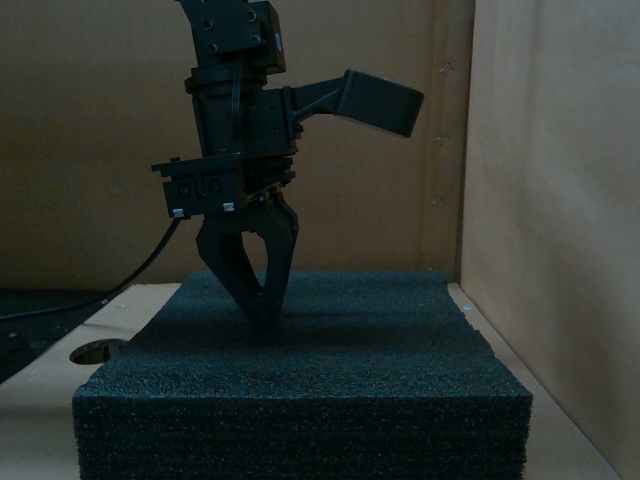}
        \caption{Base Height}
        \label{fig:foam_b}
    \end{subfigure}
    \hfill
    \begin{subfigure}[b]{0.24\textwidth}
        \centering
        \includegraphics[width=\textwidth]{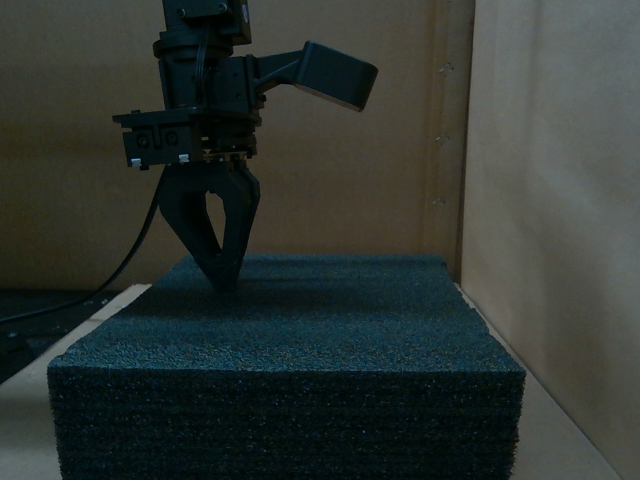}
        \caption{Raised Height}
        \label{fig:foam_c}
    \end{subfigure}
    \hfill
    \begin{subfigure}[b]{0.24\textwidth}
        \centering
        \includegraphics[width=\textwidth]{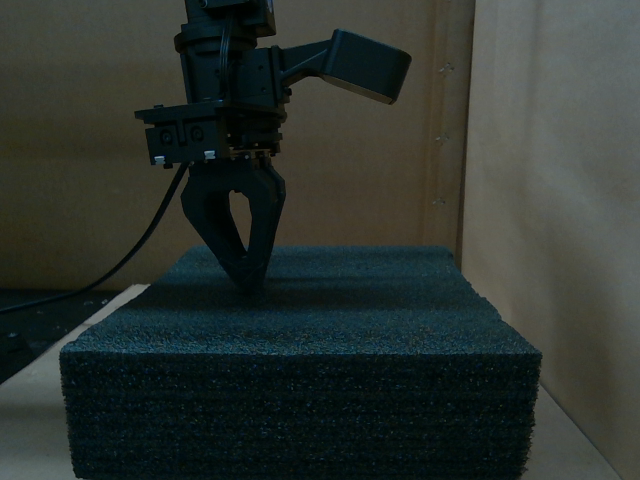}
        \caption{Max height}
        \label{fig:foam_d}
    \end{subfigure}
    \caption{\textbf{Foam Stop} The end-effector approaches a black block of foam with little textural cue to provide distance. When the End Effector creates a small amount of contact with the foam, the demonstrator stops the gripper. We create three different foam height modes, and we investigate if leveraging torque signals could help guide the robot in learning modal behavior that elicits similar equilibrium tendencies to the demonstrations (Fig. \ref{fig:torque_plot}).
    }
    \label{fig:foam_press_main}

\end{figure*}


\subsection{Empirical and Qualitative Evaluation}
\subsubsection{Board Wipe}
From Table \ref{tab:performance_table}, \textbf{ACT-o}'s 15\% success rate is reflected by its inability to maintain downward normal force $F_x$ (forward in EEF frame), leading to "hovering" behavior. While \textbf{Base ACT} achieves 60\% success by implicitly capturing force signals, the transition to explicit screw torque grounding enables 90–95\% success rates. This success demonstrates that using effort as a torque proxy provides the necessary grounding to modulate vertical motion, ensuring the agent prioritizes surface pressure even during high-velocity lateral wiping.

\subsubsection{Soft Bottle Press}
\textbf{ACT-o} exhibits underwhelming ability in generating coherent trajectories. We hypothesize this stems from a common phenomenon in ACT -- that states within vertical downward and exit trajectories are proprioceptively and visually similar; without Leader-Follower or explicit force feedback, the policy cannot differentiate between these states. While \textbf{Base ACT}'s inability to model stiffness modes results in overfitting to the low-stiffness policy mode. Conversely, incorporating screw-torque proxies to either model produce successful trajectories for either stiffness, showing that torque provides a critical auxiliary grounding necessary to resolve state ambiguity across different stiffness regimes.

\subsubsection{Plug Insert}
The addition of torque-proxies significantly enhances both the overall success and the corrective capabilities. Both \textbf{ACT+$\tau$} and \textbf{ACT-o+$\tau$} demonstrate successful rollouts characterized by a series of "adjustment-taps" that often span two episodes to reach full fruition, qualitatively learning the gentle, reactive motions necessary for alignment. In contrast, \textbf{Base ACT} exhibits stiff adjustment behaviors and a problematic failure mode; despite aligning the plug pose, it either fails to apply sufficient downward force or completely freezes upon contact. Furthermore, \textbf{ACT-o} sometimes produces invalid trajectories marked by "hesitant" behavior, wherein the policy remains stationary and fails to initiate the required task dynamics.

\subsubsection{Foam Stop}

From Fig. 4, All models were able to stop at approximately the required foam heights without pushing through or "floating". However, \ref{fig:torque_plot}, \textbf{ACT+$\tau$} and \textbf{ACT-o+$\tau$} demonstrate superior convergence toward the mean demonstration $F_x$ equilibrium value, closely matching its effort profile. These results indicate that while kinematic imitation alone is competitive, integrating a screw-torque proxy provides computationally inexpensive yet powerful physical grounding for torque-aware object interaction.

\section{Conclusion}

Our investigation reveals that the efficacy of ACT is deeply tethered to implicit interaction cues generated by tracking discrepancies in leader-follower teleoperation. By introducing the observation-centric \textbf{ACT-o} variant to isolate these cues, we demonstrate that removing the implicit signal gap causes a catastrophic collapse in performance for contact-rich tasks, and reduced general policy learning. This failure manifests as "hesitant" or "stale" behaviors where the model, deprived of the latent force-awareness typically provided by the leader arm, fails to initiate task dynamics or differentiate between visually similar states— reaching 0\% success in some task settings.

To bridge this gap, we showed that integrating explicit screw-torque proxies derived from motor current serves as a powerful and computationally inexpensive substitute for teleoperation-induced cues. These torque-aware configurations effectively restore contact-aware behavior, allowing the model to trend toward a stable dynamical equilibrium that closely matches human demonstrations in force-sensitive tasks. Beyond mere recovery, the addition of explicit torque proxies enhances fine-grained actions on top of the base policy, enabling emergent "adjustment-taps" and consistent normal force maintenance that kinematic imitation alone cannot achieve. Our findings establish a practical methodology for enabling robust, force-aware manipulation on platforms lacking specialized hardware or external force/torque sensors.

\section{Limitations and Future Work}

Our study primarily utilizes joint-side torque estimates and motor current proxies. While effective for macroscopic tasks like soft bottle pressing, these signals lack the high-frequency tactile resolution provided by dedicated six-axis F/T sensors or optical tactile arrays. Furthermore, our study prioritizes demonstrating force-aware behaviors that emerge within affordable, low-data regimes, leaving the impact of large-scale data scaling on the utility and stability of these proxies as a subject for future investigation.

We intend to further our investigation in cross-platform proxy transfer, exploring if normalized torque proxies can act as a universal language for contact dynamics across heterogeneous hardware. We also intend to investigate force-aware distillation, utilizing a VLM to generate high-level task plans while a force-integrated ACT head executes the final, contact-rich centimeter of the trajectory. Lastly, we aim to move beyond reactive policies toward active sensing, where the policy explicitly chooses to probe or tap an object to disambiguate visual uncertainty through haptic confirmation before proceeding with a manipulation sequence.






\bibliographystyle{IEEEtran} 
\bibliography{references}    

\end{document}